# Roots and Requirements for Collaborative AIs

Mark Stefik

Version April 22, 2024

The vision of AI collaborators is a staple of mythology and science fiction, where artificial agents with special talents assist human partners and teams. In this dream, sophisticated AIs understand nuances of collaboration and human communication. The AI as collaborator dream is different from computer tools that augment human intelligence (IA) or intermediate human collaboration. Those tools have their roots in the 1960s and helped to drive an information technology revolution. They can be useful but they are not intelligent and do not collaborate as effectively as skilled people.

With the increase of hybrid and remote work since the COVID pandemic, the benefits and requirements for better coordination, collaboration, and communication are becoming hot topics in the workplace. Employers and workers face choices and trade-offs as they negotiate the options for working from home versus working at the office. Many factors such as the high costs of homes near employers are impeding a mass return to the office.

Government advisory groups and leaders in AI have advocated for years that AIs should be transparent and effective collaborators. Nonetheless, robust AIs that collaborate like talented people remain out of reach. Are AI teammates part of a solution? How artificially intelligent (AI) could and should they be?

This position paper reviews the arc of technology and public calls for human-machine teaming. It draws on earlier research in psychology and the social sciences about what human-like collaboration requires. This paper sets a context for a second science-driven paper that advocates a radical shift in technology and methodology for creating resilient, intelligent, and human-compatible AIs (Stefik & Price, 2023). The aspirational goal is that such AIs would learn, share what they learn, and collaborate to achieve high capabilities.

## CONTENTS

Prologue

Over the past few years, there have been multiple high-level calls for AIs that are human compatible and that collaborate with people. However, the research projects and government-funded programs typically are about creating AIs that learn but don't collaborate or create AIs to support human-computer teams but don't learn. AI systems were not expected to collaborate and collaboration systems were not expected to learn.

This investigation for this position paper began with what John Seely Brown characterizes as "radical innovation" – multi-disciplinary research that follows a challenging problem to its root. It started with an investigation into what collaboration requires and what future AI systems that collaborate would require. Instead of sticking with one approach or perspective, radical research pivots when it gets stuck. It brings in people with different perspectives and insights to find a way forward.

Over the course of a year, the journey required additional perspectives from fields not anticipated at the beginning – including social psychology, cognitive and developmental psychology, neurosciences, AI, and many subfields such as developmental robotics and large language models.

This is the first of two papers from this work. The second paper (Stefik & Price, 2023) is about a bootstrapping approach for creating AIs and AI collaborators. This first paper focuses on collaboration per se. It draws on perspectives from social psychology and AI about what that means.

# 1 Introduction

## 1.1 Imagining AIs that Collaborate (or Don't)

People have imagined creating artificial beings and written about them for thousands of years (Truitt, 2021; Truitt, 2015). In Victorian times, mechanical automatons were created for court audiences and public amusement. Examples from contemporary science fiction include J.A.R.V.I.S. in the Iron Man movies Marvel comics, Data in Star Trek, the many droids of the *Star Wars* movies and series, Bishop and Ash in the *Alien* movies, and R. Daneel Olivaw in Isaac Asimov's books. Typically, these fictional AIs are attuned to nuances of human interactions, anticipate human needs, and can collaborate like humans.

Are AI collaborators and teams just around the corner? In *Harvard Business Review* Wilson and Daugherty urged corporations to prepare for human-AI teaming (Wilson & Daugherty, 2018). They say that human-AI teams can now boost human analytic and decision-making abilities, heighten human creativity, and enable new and powerful ways to interact with customers. The authors offer principles about how to deploy human-AI teams successfully. A contrasting business management article in *Sloan Management Review* is cautionary (Saenz, M. J. et al., 2020). It advises that today's collaborative AIs fall far short of user expectations. It suggests that different approaches are needed for different levels of risk and autonomy.

To create a hands-on experience for the public, Autodesk and the Smithsonian created the "Co-Lab" exhibit in the museum's Arts and Industries building in Washington, D.C. (Smith, 2021; Glocomp marketing, 2021). The exhibit invited the public to explore human-machine teaming. Museum goers could form teams with friends, strangers, and an AI. They could select and place virtual "tiles" to design blocks in a sustainable future city. The participants take different roles such as developer, ecologist, and mayor. They work together and have goals corresponding to their roles. The AI on the team has the role of a design generator and analyst.

Autonomous AIs are increasingly used in many settings. However, today's AIs are not usually human-compatible (Russell, 2019), become confused and fail in new situations, and do not collaborate well. Looking ahead, if AIs cannot collaborate with their users and other AIs, people will continue to distrust AIs and it will be potentially dangerous to rely on them. Many writers (e.g., Newell, 1976; Shrage, 1993; Stefik, 1999) described dystopian scenarios of such non-collaborative AIs and automation especially when unanticipated events and breakdowns occur.

What makes a good collaborator? It is easy to underestimate the skills and knowledge required for good collaboration. Human skills for collaboration are developed early in life. Newborn children do not collaborate. However, over the first two to four years of life, they learn incrementally how to collaborate. They learn how to collaborate while they are learning many other things including how to move, how the world works, what other people do, and how to communicate. In adults, skills for collaboration are ingrained and often go unnoticed until they break down.

Current AI systems can drive cars, but users are advised to have a human driver behind the wheel to take over quickly when something goes wrong[1]. Automobile navigation programs give directions, but they get confused and give nonsensical suggestions on multi-level highways and in urban canyons. Music recommendation systems advise about new music, but they don't learn properly from user feedback ("No more Beach Boys music!"). These AI systems are tools that can augment human driving, way finding, and music finding. But they do not collaborate well.

## 1.2 A First Wave of Computer-Intermediated Teaming and Human Augmentation

J.C.R. Licklider and Robert Taylor proposed the idea of "human-machine symbiosis" to improve the performance of human teams in understanding complex situations and make plans (Licklider, 1960; Licklider and Taylor, 1968). At that time, the predominant way of interacting with computers was by typing and viewing text and graphics on low resolution displays. Computers carried out predetermined processing on pre-formulated problems. Licklider and Taylor recognized the potential for collaborative teams to use computers in meetings to retrieve information to support their deliberations, to focus attention by pointing to displayed information, and to create runnable models that would transcend human limitations and avoid mistakes. Computers would find computational solutions to problems, and humans would set goals, formulate hypotheses, and evaluate findings and solutions.

Trained in psychoacoustics, Licklider became interested in information technology. He joined the Defense Advanced Research Projects Agency (DARPA) where he founded the Information Processing Techniques Office (IPTO). IPTO funded research at computer science departments and research centers, including the Augmentation Research Center at SRI (Engelbart, 1962) headed by Douglas Engelbart. Robert Taylor was a program manager at NASA's Office of Advanced Research and Technology. Taylor had done research as a graduate student in neuroscience and psychoacoustics. He became a strong advocate of Licklider's vision and later held the same role as head of IPTO.

In *Man-Computer Symbiosis* Licklider used a concrete example to motivate intelligence augmentation (IA). It was based on an informal time-and-motion analysis of his activity working on a technical problem.

---

[1] See Eric Vinkhuyzen's video blog (Vinkhuyzen, 2022) for a perceptive analysis of how self-driving cars are not yet competent and why they create confusion for other drivers and pedestrians. The problem is not just the long tail of edge cases of possible traffic situations. Driving is a social activity, not a solitary one. Automatic drivers do not understand and do not participate in the social signaling that supports safe and cooperative driving. Human drivers are expected to understand signals of intention that make actions predictable, coordinate smooth interactions with others, and support mutual safety. Currently, automatic drivers cannot read social situations and do not coordinate with other parties on the roads and sidewalks.

> "About 85 per cent of my "thinking" time was spent getting into a position to think, to make a decision, to learn something I needed to know. Much more time went into finding or obtaining information than into digesting it. Hours went into the plotting of graphs, and other hours into instructing an assistant how to plot. When the graphs were finished, the relations were obvious at once, but the plotting had to be done in order to make them so. … When they were in comparable form, it took only a few seconds to determine what I needed to know." (Licklider, 1960)

Licklider's account anticipates later studies (e.g., Pirolli & Card, 2005; Russell et al., 1993) of the sensemaking practices of career intelligence analysts and others. These cognitive analyses showed how people make sense of situations as they gather and organize information. They create external models of the situations and negotiate interpretations and conclusions.

Licklider's long-term vision was that people and computers would work closely together, each focusing on the parts of a task where they are most capable. Together they would formulate goals and find solutions. His preliminary analysis suggested that human-computer partnerships would perform intellectual operations better than people or computers alone. Recognizing that his vision of human-computer symbiosis was ahead of its time, Licklider called out prerequisites. These included improving computer hardware and memory and improving interactive displays with wall-sized displays for teams. In Licklider's view, the greatest obstacle to human-machine symbiosis was the inability of computers to communicate in human language.

Licklider and Taylor focused on the development and sharing of mental models. The models are the conceptual structures and abstractions that describe what the participants are communicating about.

> "By far the most numerous, most sophisticated, and most important models are those that reside in men's minds. In richness, plasticity, facility, and economy, the mental model has no peer, but, in other respects, it has shortcomings. It will not stand still for careful study. It cannot be made to repeat a run. No one knows just how it works. It serves its owner's hopes more faithfully than it serves reason. … It can be observed and manipulated only by one person." (Licklider & Taylor, 1968)

In collaborative sensemaking, each person collects and analyzes information and assembles and synthesizes it. Collaborative model building is central to progress in the intellectual activity.

> "Society rightly distrusts the modeling done by a single mind. Society demands consensus, agreement, at least majority. Fundamentally, this amounts to the requirement that individual models be compared and brought into some degree of accord. The requirement is for communication, which we now define concisely as "cooperative modeling" – cooperation in the construction, maintenance, and use of a model. …
>
> We can say with genuine and strong conviction that a particular form of digital computer organization, with its programs and its data, constitutes the dynamic, moldable medium that can revolutionize the art of modeling and that in so doing can improve the effectiveness of communication among people so much as perhaps to revolutionize that also." (Licklider & Taylor, 1968)

Licklider and Taylor contrasted their understanding of human communication with how a communications engineer might describe two tape recorders sending and receiving information to and from each other. In their view, such rote transfer of information had little to do with how people communicate. Using the computer as a communications medium, team members perceive an external representation. The representation could be spoken, written, or presented in drawings and figures. They manipulate the representation and consider the evidence and its provenance. They compare their own internal mental models to the shared external models. In their collaborative modeling, members adjust their own understandings and incrementally and refine the externalized representations.

Since then, psycholinguists (e.g., Clark & Brennan, 1991) have studied how people create and maintain common ground in discourse. Multidisciplinary studies have analyzed socially shared cognition (e.g., Resnick et al., 1996) at various timescales. For Licklider and Taylor, the main point was that computers were well suited to be an interactive medium for supporting human teamwork. They believed that interactive computers could be far more powerful aids to collaboration than blackboards in meetings and published documentation.

Building on the personal Dorado computers and local networking technology created by Taylor's Computer Science Lab, PARC researchers (Stefik et al., 1987) created an electronic meeting room (the "Colab") and team collaboration software that had multi-user interfaces for organizing ideas, telepointers, coordination signals, and an interactive wall-sized display.[2] In *Toward Portable Ideas* the authors extended the vision of meeting support and described how persistent computer-augmented conversations could span an organization (Stefik & Brown, 1989). The Colab and clones of it were used at PARC and other Xerox sites for several years until the PC revolution obsoleted their early personal computers. During the COVID pandemic, millions of people experienced computer-intermediated meetings as organizations incorporated more remote work.

More ambitious than today's remote meeting tools, Licklider and Taylor envisioned computational and *interactive modeling facilities for communities* where computers would support automated reasoning, dynamic visualization, and interactive exploration.

A decade after Licklider and Taylor's paper, Joshua Lederberg reflected on five years of experience in creating and using AI and digital communication technology in the SUMEX medical and scientific research community (Lederberg, 1976). He described how technology was changing the conduct of science including the testing and refining of scientific findings using rich, knowledge-based digital models. Today, executable digital models (sometimes called "digital twins") are created for advanced engineering and scientific applications. These state-of-the-art systems are useful, but even for experts they are challenging to understand, use, and maintain. They are sophisticated tools, but they do not collaborate like people.

Mitch Waldrop (Waldrop, 2001) and Walter Isaacson (Isaacson, 2014) credit Licklider's and Taylor's leadership and vision of augmented human intelligence as the driving forces of the computer technology revolution.[3] It set in motion the advances in personal computing, graphical user interfaces, the internet, online communities, and other key elements of computing today. More recently, AI systems processing information from the internet, have taken the revolution and its social impact to new levels.

Stepping back, the focus of intelligence augmentation is on computers that help people to solve problems better together. The focus of AI is on intelligent computers that use intelligence autonomously to do things for people. In the context of AI, there is a significant shift in the role of computers. In IA, the computers are tools that provide effective infrastructure for human collaboration. In AI, computers are elevated from collaboration infrastructure to teammates.

In his history of Silicon Valley, *Machines of loving grace: The quest for common ground between humans and robots,* John Markov builds on this vision (Markov, 2015). He suggests that the combined themes of IA (intelligence augmentation) focusing on people and AI (artificial intelligence) focusing on intelligent machines is compelling for the next phase of the computer revolution. Creating embodied collaborative, and trustworthy AI partners for homes, all manner of services and businesses, and even tending the environment could have enormous benefits for improving human life and potential.

---

[2] The Colab project was funded by DARPA contract N00140-87-C-9258.

[3] Other accounts of the revolution focus on the advances in semiconductors that enabled creating computers with high performance and low cost. In these stories, AI and IA speak to what the computers are for.

## 2 What Competencies are Needed for Collaboration?

What do AIs need to be good collaborators? The maxim "many hands make light work" conveys the wisdom that people do better working together than working alone. What does collaboration require?

In 2004 Gary Klein and his colleagues reflected on what automated systems need to collaborate like people do.[4] They focused on tight knit teams working on shared tasks and published their findings in *Ten Challenges for Making Automation a Team Player* (Klein et al., 2004). The paper's ten challenges are listed in Table 1 in abbreviated form.

1. Maintaining common ground
2. Modeling task and teammates
3. Predicting teammate actions and being predictable
4. Being directable
5. Making current state obvious
6. Communicating status and intention
7. Negotiating goals
8. Collaborative planning
9. Managing attention
10. Managing costs of coordination

*Table 1. Ten challenges for human-AI teaming. Adopted from (Klein et al., 2004)*

The authors of the "Ten Challenges" paper assessed that the state of the art for AI systems fell short of meeting the challenges. It still does. In the following we draw on these ten challenges to illuminate properties of collaboration as practiced by people and to inspire competencies and research milestones relevant to creating AIs whose behavior is more like the behavior of human collaborators.

---

[4] Klein and his colleagues did not advocate that human-like collaborators should be created. Their paper conveys the deep challenges that would need to be faced to meet that challenge. Victoria Groom and Clifford Nass, among others, take up the issue of whether computers could or should be teammates (Groom & Nass, 2007). They suggest that researchers who are interested in this idea are overly optimistic and have crucial blind spots. They suggest that if computers cannot have the required properties, then the concept of "computer teammates" should be replaced by a more accurate and appropriate model where a robot is understood as a tool. They correctly say that "robots are currently unable to meet human's high expectations for appropriate team behavior in the unpredictable, high-stakes situations for which they are designed." Their paper provides further examples of challenges in the spirit of those in (Klein et al., 2004), particularly regarding issues in the earning and negotiating of trust.

## 2.1 Maintaining Common Ground

The first challenge says that there must be agreement to collaborate. The participants intend to work together to facilitate coordinated action, work toward shared goals, and repair breakdowns.

The first challenge is as follows:

> "To be a team player an intelligent agent must fulfill the requirements of a basic compact to engage in common-grounding activities." (Klein et al., 2004)

The authors explained the meaning of common ground,

> "The Basic Compact is not a once-and-for-all prerequisite to be satisfied but rather has to be continuously reinforced or renewed. It includes an expectation that the parties will repair faulty knowledge, beliefs, and assumptions when these are detected. Part of achieving coordination is investing in those actions that enhance the Compact's integrity as well as being sensitive to and counteracting those factors that could degrade it."

In the following we use short scenarios to illustrate collaboration challenges. Consider a scene where two people are cleaning two adjacent rooms.

> In the first room, Bob prepares to clean the floor. He plugs in the vacuum cleaner and moves furniture away from the walls. A phone rings. Bob leaves for a moment to answer it. In the second room, Steve is picking up items from the floor. He notices the vacuum sitting in the first room. He unplugs the vacuum and moves it to the second room.

Neither Bob nor Steve is paying much attention to what the other person is doing. Because Bob and Steve interfere with each other's work, it takes them more time to clean the two rooms. In the scenario they do not have a shared plan and do not benefit from each other's work. With coordination, they could do the entire task more quickly. For example, two people can move heavy furniture more quickly than one.

Imagine the following conversation where Bob and Steve plan their collaboration.

> **Bob**: "There would be less wear on the rugs if we carried the furniture together rather than sliding it. Then one of us could vacuum while the other cleans the windows and dusts the surfaces."
>
> **Steve**: "That works. Then we could move the furniture back so the middle of the room could be vacuumed. One of us could use the vacuum in the next room while the other empties the waste baskets."
>
> **Bob**: "OK. The vacuum is in the other room. Shall we work on that room first?"

Young children demonstrate planning abilities activities readily plan in new and diverse contexts – whether they are passing bowls of food during a meal, folding laundry, putting it away, cleaning a room together, or building a tower of blocks. A challenge for collaborative AI is understanding how they can do this without any formal training.

The first challenge says that collaborators agree to repair common ground when there is a misunderstanding. Consider the following scenario where a service technician for wood stoves has had several phone calls with a customer trying to diagnose a creosote issue.[5]

> **Customer:** It happened again this morning. Another stream of sticky black fluid is dripping into the wood stove."
>
> **Technician:** "You are not burning eucalyptus or pine, and the wood is seasoned, right?"
>
> **Customer:** "Right."

---

[5] This scene was adapted from a story in Busha and Morris (1982). The book is an amusing guide to the history, design, maintenance, diagnosis, trials, tribulations, and joys of having a wood stove. The book condenses lessons from "Stove School." It is used for factory training of representatives of Vermont Castings, Inc.

> **Technician:** "We discussed the stove's installation yesterday. Will you be home tomorrow morning, so that I can stop by?"
>
> The next morning, the technician inspects the wood stove.
>
> **Technician:** "What is that thing attached to the damper?"
>
> **Customer**: "That's a five-ounce fishing weight."
>
> **Technician** (frustrated): "Why?"
>
> **Customer**: "It was my wife's idea. The damper kept clinking every time it closed."
>
> **Technician**: "Why didn't you tell me that you put a five-ounce lead weight on your air inlet damper?"
>
> **Customer**: "You never asked."

In this scenario the technician has a detailed understanding about how wood stoves and chimneys function. The customer lacks this understanding. He just wants a cozy fire in a quiet room.

The technician did not imagine that the customer would alter a wood stove in a way that interferes with its operation. Each party expects the other to inform them about everything that they need to know. In this example, neither party understood what the other party needed to know. The technician was surprised when he discovered that salient facts had not been disclosed. This is where maintaining and repairing common ground comes in.

The next scenario is about a situation in an open world with many unknowns. This fictional conversation about a ship's movements takes place between a human analyst and an AI assistant.

> **Analyst**: "The ship arrived late. They must have been running the engine slower than usual."
>
> **AI**: "Not necessarily. There might have been heavy head winds or floating debris in the region. However, nothing like that was recorded in the communications log."
>
> **Analyst**: "Fair points. I notice that the ship was near a conflict zone. It would need to maintain radio silence. Also, there are protocols about what should not be recorded in conflict situations. Some data may have been omitted from the communications log. What other data sources did you consult?
>
> **AI**: "Now checking more sources. No windy weather or debris obstacles are reported in the weather channel. However, there is an anomaly in the ship's wake according to image data from satellite delta 8. The ship went off course and stopped for thirty minutes. It may have rendezvoused with another ship to transfer troops or equipment. No wake is visible from a second ship, but it could have been a submarine."
>
> **Analyst**: OK. Let's flag this analysis as sensitive and send it up the chain.

Each party in this scenario has a cognitive model. They use their own models to interpret the data, but they lack full knowledge of each other's models.

The analyst made a routine analysis based on the assumption that the case was typical. Extending common ground, the AI demonstrates its understanding that wind or obstacles could slow a ship's movement. It points out that the analysis did not consider those possibilities. The analyst then reconsidered and mentioned that sensitive data might be missing from the communications log because of reporting policies near a conflict zone. When the AI inspected image data from a non-military satellite, it found a navigational anomaly from image and timing data of the ship's wake. The AI then extended its interpretation of the ship's delayed arrival time based on that context.

In summary, effective collaboration requires that collaborators maintain a shared understanding about what they are doing. They watch for and address breakdowns.

## 2.2 Modelling Tasks and Teammates

The second challenge is as follows:

> "To be an effective team player, intelligent agents must be able to adequately model the other participants' intentions and actions vis-à-vis the joint activity's state and evolution— for example, are they having trouble? Are they on a standard path and proceeding smoothly? What impasses have arisen? How have others adapted to disruptions to the plan?"

Challenge 2 stipulates that each collaborator has models of the others and the tasks that they are doing. Consider the following example.[6]

> Macer is a toddler. He is on the floor building a tower of blocks. He sees his dad (Nick) tape a ribbon to the last wrapped package on the table. Nick puts the ribbon, scotch tape, and wrapping paper in a box.
>
> Gathering up the packages, Nick walks to the hall closet. He plans to hide the presents in the closet so that Macer's mom will not see them before her birthday celebration that night.
>
> Standing by the closet door, Nick struggles to hold all the packages using one arm. He looks at Macer.
>
> Macer gets up from playing with his blocks on the floor. He runs to his dad. Macer turns the doorknob, opens the closet door, and holds it open.
>
> Nick puts the packages on a shelf in the closet. He smiles at Macer and gives him a hug.

This scenario illustrates a young child meeting challenge 2. Macer notices that his dad has his hands full. He opens the closet door so that his dad can use both hands to put the birthday presents on a closet shelf. Parents of kids as young as two or three observe them helping like this.

People have models of everyday activities, such as eating, preparing meals, getting things out of cupboards and closets, stacking things in piles, carrying things, and opening and closing doors. People use models to make predictions about what happens next in activities. They develop models of causes and effects. For example, when people put food in their mouths, they predict that it will have a certain taste. When a person picks an object up, the object is not where it was before. A ribbon taped to a package moves with the package when the package is moved.

Children develop models for things that they carry. When his dad puts items in a box, Macer knows that they will disappear from his view. He knows that if he loosens his grip on an object that he is carrying, it will fall. Models can be combined. In the scenario, Macer knows that his dad needs to use both hands to carry the presents. He predicts that his dad will use one hand to turn the knob and open the closet door. He knows that his dad does not want to drop the packages. He predicts that if his dad has only his other hand to hold the packages, some will fall to the floor. Macer predicts that if he helps his dad, he may get a smile and a hug afterwards.

In summary, to collaborate effectively, collaborators need models of tasks. They use models to make predictions. They use the predictions to help them decide what to do and to detect surprises based on misunderstandings. In current engineering practice, developers need to create models of tasks for every new application.

---

[6] This scenario is inspired by findings and scenarios of helping by human children (Warneken & Tomasello, 2006) and by online videos of similar scenarios (PBS, 2012; Tenenbaum, 2019).

## 2.3 Predicting Teammates and Being Predictable

Challenge 3 says that human-AI team members must be mutually predictable. Without predictability, people cannot anticipate what their collaborators will do next, and collaboration breaks down. Similarly, a collaborative AI should predict the actions of others and act in ways that enhance the ability of others to predict its actions. The challenges paper says:

> "… To be a team player, an intelligent agent—like a human—must be reasonably predictable and reasonably able to predict others' actions. It should act neither capriciously nor unobservably, and it should be able to observe and correctly predict its teammates future behavior."

Adversarial activities such as team sports provide nuanced considerations about predictability. Football players want to be predictable to members of their own team and unpredictable to members of the opposing team. Each team uses hand signals, body language, secrets, and misdirection to coordinate its own members and to mislead the other team.

Consider the following fictional scenario about a football play.

> The blue and red teams are at the line of scrimmage.
> The teams poise for action.
> The blue quarterback recites a sequence of numbers: "Sixteen, four, seven, two, forty-three, eight …"
> The blue center suddenly hikes the football to the quarterback.
> As a red tackle approaches the blue quarterback, the quarterback pivots and passes the ball laterally to a blue halfback who is already running.
> The blue running back races down the field.

In this scenario, the quarterback and the blue team share secrets about the numbers that signal the start of a play and what play will be executed. Both teams will jump into action the moment the ball is hiked, but only the blue team knows which numeric pattern signals the starting moment. This numeric version of "ready, set, go" cues the blue team just before the main action begins.

Each team prepares for games in their pre-game practices. In this scenario, a feint to the blue tight end is intended to misdirect attention of the red team away from a blue halfback. Pre-game practice helps the teams to get their timing right. Plays are designed based on the comparative strengths of the two teams and refined after during practice.

All the scenarios involve communication. Besides speech, people often use signals to coordinate and influence the actions of others.

## 2.4 Seven More Challenges

The seven remaining challenges are listed in Figure 1.

> 4. Being directable.
> 5. Making current state obvious.
> 6. Communicating status and intention.
> 7. Negotiating goals.
> 8. Collaborative planning.
> 9. Managing attention.
> 10. Managing costs of coordination.

*Figure 1. Seven more challenges to teaming.*

Like the first three challenges, challenges 4 through 6 are about communication, coordination, and models that enable them. Collaborators model each other and act in ways that help other collaborators know what they will do next.

Challenges 7 and 8 introduce concepts where a team's aggregate behavior must be considered rather than just the behavior of one individual. Teams have collective goals. Negotiation in team planning must consider multiple collaborators, multiple goals, and tradeoffs. Negotiation requires models of conversation, turn taking, and multiple agent planning.

Challenges 9 and 10 reflect tradeoffs between coordination and efficiency. Teams may skip some centralized planning if the costs of planning are higher than the savings from coordination.

The ten challenges describe behaviors and competencies that are characteristic of adult collaboration. Tests based on such challenges can be used in evaluating AIs as collaborators (Kozierok et al., 2021a-b).[7]

---

[7] In their role as the test and evaluation team Kozierok et al. published the report *Hallmarks for Machine Collaboration* DARPA's "Communicating with Computers" (CwC) program. Their goal was to evaluate machine communications on open-ended complex activities where the machines are collaborating with people. Their tests were not specifically designed to test competencies of the collaborating AIs, but rather to examine properties of the communication acts between the humans and the machines. Nonetheless, the categories in their "hallmarks" were closely related to the ten challenges in the Klein et al. paper. For example, they included detection and repair of misunderstandings (maintaining common ground) and context awareness (making current state obvious) in their categories of communication.

## 3 The Research Trajectory of Collaborative AI

In the following we draw from AAAI presidential addresses to represent how the AI research community's vision of collaborative AIs has evolved as the challenges of collaboration have become more deeply understood.

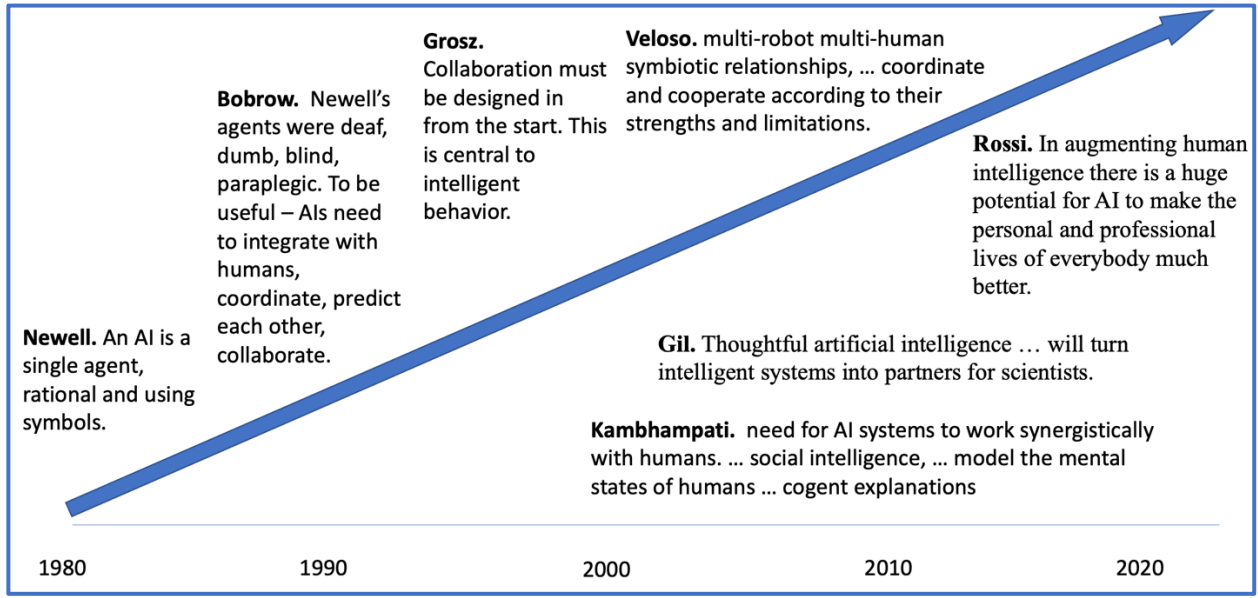

*Figure 2. AAAI Presidential visions of human-AI collaboration*

In 1980 the first president of AAAI, Allen Newell, summarized his view of AI in terms of a standard view and open issues. He described knowledge versus symbol levels of analysis, a principle of rationality, and the role of logic in creating AIs (Newell, 1981). A decade later, Daniel Bobrow in his 1990 AAAI presidential address (Bobrow, 1991) observed that Newell had characterized AI systems as individual, rather isolated agents. This was not to suggest that Newell did not think about humans and computers. In *Human Problem Solving* (Newell & Simon, 1972) Newell developed step-by-step protocols of humans solving problems as guides for computers to solve problems like people do. In *The Psychology of Human-Computer Interaction* (Card, et al., 1983) Newell together with Stuart Card and Thomas Moran put analytic foundations under the design of human-computer interactions, helping to establish the field of information processing psychology. In *Unified Theories of Cognition* (Newell, 1990), Newell described "The Social Band" as a frontier of modern psychology and social psychology for analyzing multi-agent situations. Nonetheless, in his AAAI address about the state and future of AI, Newell did not include AIs that collaborate with humans and other agents. Bobrow observed,

> "Newell's analysis postulated a single agent, with fixed knowledge and a specified goal. The agent was disconnected from the world with neither sensors nor effectors, and more importantly with no connection to other intelligent goal-driven agents. Research results in AI consisted primarily in the determination of the principles of construction of such intelligent, but deaf, blind, and paraplegic agents. … there has been increasing recognition of the importance of designing and analyzing system that do not make these isolation assumptions." (Bobrow, 1991)

Bobrow drew on examples from multiple potential applications. He called out communication, coordination, and integration as essential types of interactions for AIs and the requirements for each. For example, to communicate AI agents need to establish common ground meanings with others. To share resources, they need to coordinate and predict each other's actions. To be useful, they need to integrate with human social practices. Stretching the vision of Licklider and Taylor, Bobrow predicted that the Internet would evolve from providing a communications medium for information to a knowledge medium (Stefik, 1986), where AI agents provide knowledge-based services. He challenged researchers in the next decade to program AI agents to interact directly with other AIs and humans in the world.[8]

In her 1996 AAAI presidential address (Grosz, 1996), Barbara Grosz called on the AI community to create "computer systems that are intelligent, collaborative problem-solving partners."

> "… the capability to collaborate with users and other systems is essential if large-scale information systems of the future are to assist users in finding the information they need and solving the problems they have. … Collaboration must be *designed into systems from the start*; it cannot be patched on. … That collaboration is *central to intelligent behavior* is clear from the ways in which it pervades daily activity. … They range from the well-coordinated, pre-planned and practiced collaborations of sports teams, dancer, and musical groups to the spontaneous collaborations of people who discover they have a problem best solved together." (Grosz, 1996)

As evidence that the time was right for addressing this challenge, Grosz cited progress on multi-agent planning, models of distributed AI based on negotiation (Davis and Smith, 1983), and research in other disciplines about computer-intermediated work. She distinguished AI systems seen as tools from AI systems seen as collaborators.

In his 2004 AAAI presidential address, Ronald Brachman reflected on why multidisciplinary approaches and system integration and evaluation are essential for AI to reach its potential. This advice is pertinent to creating collaborative AIs. Earlier in his career Brachman served as office director of IPTO, the DARPA office started by Licklider. He launched a Cognitive Systems Program at DARPA.

In his 2008 AAAI presidential address, Eric Horvitz included in his core research challenges for AI the creating of common ground for humans and AIs about shared references, beliefs, and intentions. His research contributions include mixed-initiative human-machine interfaces and what he called complementary computing which is about coordinating AI systems.

In 2012 the incoming AAAI president, Manuela Veloso, published three papers on robotics at the AAAI conference. She also co-authored thirty-one papers with students and colleagues that year in other journals and conferences on neural information processing, robotics, autonomous systems, multi-agent systems, human-robot communication, and multi-agent planning. The goal of her research group at Carnegie Mellon University was to enable multi-robot multi-human symbiotic relationships, in which humans and robots coordinate and cooperate according to their strengths and limitations. Veloso and her team are established leaders in designing and building robots, robot teams, RoboCup soccer, and cobot service robots doing joint tasks with people. Veloso subsequently authored papers where robots and people help each other in carrying out tasks. Research toward collaborative AI was being published in many venues.

---

[8] Bobrow's view was in line with that of those researchers in AI, artificial life, and developmental psychology that viewed the coupling of the body, brain, and environment as essential for the development of cognition. Newell's presidential address was the mainstream view for AI and cognitive science at the time. His focus was on the rule-based manipulation of symbols by the symbol processing machinery of an agent. A decade after Bobrow's address, Lungarella and colleagues published a review of this history and of the beginnings of developmental robotics (Lungarella et al., 2003).

In his 2016 AAAI presidential address, Subbarao Kambhampati focused on "making AI systems interact, team, and collaborate with humans" (Kambhampati, 2020).

> "[A]s AI technologies enter our everyday lives at an ever-increasing pace, there is a greater need for AI systems to work synergistically with humans. To do this effectively, AI systems must pay more attention to aspects of intelligence that help humans work with each other — including social intelligence, … modeling the mental states of humans-in-the-loop and recognizing their desires and intentions, providing proactive support, exhibiting explicable behavior, giving cogent explanations on demand, and engendering trust … [T]he quest for human-aware AI systems broadens the scope of AI enterprise; necessitates and facilitates true interdisciplinary collaborations; and can go a long way toward increasing public acceptance of AI technologies. … Making such human-aware AI agents, however, poses several foundational research challenges that go beyond simply adding user interfaces ..." (Kambhampati, 2020)

Kambhampati's research with colleagues at Arizona State University included addressing fundamental problems motivated by challenges of human-aware AI. His research studies decision making, multi-agent planning, and AI architectures for explaining and reconciling models.

By 2020, there had been significant progress creating AIs to augment human abilities and to support sharing of external models. In her 2020 AAAI presidential address, Yolanda Gil focused on the challenges of AI partnering in scientific research. She reflected on one of AI's grand challenges, where AIs carry out scientific discovery and as she cast it, write scientific papers (Gil, 2021). Her address draws on her earlier paper (Gil, 2017) that envisioned AIs as full scientific partners.

> "While in recent years computers have propelled science by crunching through data and leading to a data science revolution, qualitatively different scientific advances will result from advanced intelligent technologies for crunching through knowledge and ideas. … Thoughtful artificial intelligence … will turn intelligent systems into partners for scientists." (Gil, 2017)

Gil focused on the activities and requirements for conducting science. Her address extends the vision of earlier programs such as the National Collaboratories originally conceived by William Wulf and others (National Research Council, 1993) and on the findings of the NSF workshop that she led on discovery informatics (Gil & Hirsh, 2012).

> "These intelligent systems should be capable of taking on significant problems by formulating their own research goals, proposing and testing hypotheses, designing theories, debating alternative options, and synthesizing new knowledge. They should be able to explain their reasoning, compare their lines of inference to other possible ones, and situate their findings. Intelligent systems should be able to communicate with scientists with different levels of expertise and understanding in a topic. To form a true partnership, they should be able to take guidance from scientists as well as provide guidance to them in turn." (Gil, 2021)

Elected AAAI president for 2022, Francesco Rossi is known for her research on constraint satisfaction techniques in multi-agent planning and scheduling, learning of solution preferences, semantics of concurrent programs, and social choice. In an invited talk, she reflected on limitations of current AIs saying that they lack adaptability, generalizability, self-control, consistency, common sense, and causal reasoning. Drawing on Kahneman's book, *Fast and Slow Thinking* (Kahneman, 2011), she proposed an architecture for advancing AI towards having such capabilities. In an interview in 2017 (Conn, 2017), Rossi described research on augmenting human intelligence and human-AI teaming.

> "I personally am for building AI systems that augment human intelligence instead of replacing human intelligence. And I think that in that space of augmenting human intelligence there really is a huge potential for AI in making the personal and professional lives of everybody much better. … [M]ore AI systems together with humans will enhance our kind of intelligence, which is complementary to the kind of intelligence that machines have, and will help us make better decisions, and live better, and solve problems that we don't know how to solve right now." (Conn, 2017)

In summary, the research vision for collaborative AI (as reflected in the inaugural presidential addresses of the leaders of the Association for Artificial Intelligence) has moved over several decades from being below the radar to strong advocacy where this direction is seen as crucial for addressing many long term challenges for research in the field and a human-centered socially-positive direction for creating AIs that are able to partner with people and advance the common good along many directions.

## 4 Who Needs Collaborative AI?

Calls for research investment in human-AI teaming are prominent in recent technology and strategy advisory plans, program announcements, and roadmaps from the U.S. Department of Homeland Security (McAleenan, 2020; Mitchell, 2021), Department of Defense (DARPA, 2022; ONR, 2022; Shyu, 2022; Ryan, 2018; Allyn, 2017) government advisory organizations (National Academies of Sciences, Engineering, and Medicine, 2022; Kozierok et al., 2021a-b; Kratsios, 2019; Schmidt et al., 2021), workshop reports on human-AI teaming research (Seeber et al., 2019; Laird et al., 2020), and critical whitepapers (Siddarth et al., 2021). Similarly, an advisory workshop on scientific machine learning (Baker, 2019) focused on the challenges of creating interactive, robust, transparent, multi-scale models of scientific phenomena. AI was central in that vision.

> "Human-AI collaborations will transform the way that science is done." (Baker, 2019)

Beyond the strategic importance of AI in national defense and strategic goals, the reports concur that human-AI teaming will increase the adoption and impact of AI applications across the board. Summarizing this theme, the *Final Report* of the National Security Commission on Artificial Intelligence says:

> "Mastering human-AI collaboration and teaming is a foundational element for future application of AI. Synergy between humans and AI holds the promise of a whole greater than the sum of its parts." (Schmidt et al., 2021)

Several recent books have been published to distinguish the hype from the reality. In the book *Rebooting AI: Building Artificial Intelligence We Can Trust* (Marcus & Davis, 2019), Gary Marcus and Ernst Davis analyze the limitations of symbolic and machine learning approaches to creating AIs from a cognitive science perspective. They suggest that new approaches are needed where AIs learn like children. In his book *Human Compatible: Artificial Intelligence and the Problem of Control* (Russell, 2019), Stuart Russell describes technical and ethical issues for AI. He assesses risks, challenges, and prospects in the perspective that AIs fundamentally need to work with people. David Mindell's book *Our Robots, Ourselves: Robotics and the Myths of Autonomy* (Mindell, 2015) is grounded in applications in extreme environments where neither fully remote operation nor full autonomy is workable. He focuses on interactions in human-machine teaming. He asks, "Where are the people and what are they doing?" Ben Shneideman's book *Human-Centered AI* (Shneideman, 2022) offers a framework for considering the meaning of intelligence, human-centered design, governance, ethical considerations, and the future of AI.

In 2016 when DeepMind's AlphaGo AI defeated the reigning Go player in China, Lee Sedol, it caught widespread public attention. This event was déjà vu of 1997 when IBM's Deep Blue defeated the world chess champion, Garry Kasparov. The attention around these news events and other recent AI milestones has reinforced the idea that AIs will automate important decisions when they get smart enough.

AI has continued to make impressive inroads in applications such as in self-driving cars, digital assistants, and other areas. In most of the recent applications the advances have been powered by deep learning (Bengio et al., 2021; Lecun et al., 2015). AI use has expanded to other applications with the goals of automation and better decisions. This has prompted concern that decision making should be transparent, and that people should oversee important decisions when there are significant risks or the possibility of unfair or illegal bias. Unfortunately, most AIs are still unable to explain their reasoning (DARPA, 2016; Gunning, et al., 2021) or modify their actions in collaboration with their users.[9] This limits their potential for adoption.

Furthermore, rather than improving system performance and saving labor on difficult tasks, autonomous systems often reduce performance and increase the training burden for users. Jack Blackhurst, Jennifer Gresham, and Morley Stone identified this issue as *The Autonomy Paradox* (Blackhurst et al., 2011). Struggling to manage autonomous systems, operators and users are left with questions that they cannot answer.

> "What is it doing? Why is it doing that? What is it going to do next? …
>
> In systems with low autonomy, like those that are completely teleoperated, the machines can easily become a burden on their human counterpart. The human serves essentially as the robot's caretaker and is unavailable for any other task while performing this role. … As self-directedness and autonomy increase, humans are freed to concentrate on higher-level tasks. But the self-directedness of the robot often exceeds its competence, and humans may trust the system more than they should." (Blackhurst et al., 2011)

David Mindell is MIT's Dibner Professor of the history of Engineering and Manufacturing and expert on human-centered robotics in the deep sea and other extreme environments. He cautions against myths of "utopian autonomy" for today's autonomous cars and more broadly for robotics in the workplace of the future:

> "What have we learned from extreme environments that might shed light on possible futures for autonomous cars? We know that driverless cars will be susceptible to all of the problems that surround people's use of automation in the [extreme] environments we have examined – system failures, variability of skills among uses, problems of attention management, the degradation of manual skills, and rising automation bias as people come to rely on automated systems.
>
> The most challenging problem for a driverless car will be the transfer of control between automation and the driver …" (Mindell, 2015)

In 2019 Matthew Johnson and Alonso Vera reflected on the myths of robot autonomy and the autonomy paradox and responded with a call to action – *No AI is an Island* (Johnson & Vera, 2019). They urged the AI research community to develop AIs that have teaming intelligence. Their advice goes beyond adding a human-aware capability to AIs.

> "The growth of sophistication in machine capabilities must go hand in hand with the growth of sophistication in human-machine interaction capabilities. Machines do not automatically

---

[9] PARC's XAI project, COGLE (Stefik et al., 2021), was funded under DARPA contract FA8650-17-C-7710.

get simpler to use because they have gotten smarter. Indeed, just the opposite is usually true. This correlation can be seen in more advanced commercial airliners, which typically require longer training times for type ratings than their predecessors did. … [Teaming] should be viewed as an approach to what AI capabilities should be built, and how, so as to imbue intelligent systems with teaming competence." (Johnson & Vera, 2019)

In an earlier paper (Johnson et al., 2014), Johnson and his colleagues analyzed the *interdependence* of collaborators on tasks. With interdependence, what happens when one collaborator does something depends on what others do. Both papers focus on designing AIs for given applications that require humans and AIs to work together to carry out a task. The design methodology identifies what parts of the task would be done by humans and what parts by an AI. They proposed a "Coactive" design methodology that analyzes task interdependence around three dimensions: observability, predictability, and directability (OPD). The OPD analysis informs design decisions for how the parties do signaling and coordination.

Since Licklider and Taylor's early papers about human-augmentation, there have been multiple waves of AI technology. Viewed from a high level, the roots of difficulty in building AIs are fundamental and reappear in each wave of AI technology. A comprehensive review of the technology waves and their limitations is beyond the scope of this paper. They are described in AI textbooks.[10]

To illustrate a fundamental issue of scale, consider the following fictional conversation among students in an early AI class from (say) the 1970s. These classes were about symbol processing languages, theorems about the limitations of computing, search, and the like. Techniques for machine learning were mentioned but not regularly practiced. The practical way to create an AI system that has intelligent behavior was to build it, that is, to specify it manually. A favorite introductory example was the use of modus ponens in English-based logic statements like in the following example. Students enjoyed thinking up similar edge cases and making fun of the prospects for building large systems based on logic. Consider the following fictional conversation.

> Student 1. "All birds can fly. That robin is a bird. Therefore, it can fly.
> Student 2. "But what if the bird's wings are clipped or damaged?"
> Student 3. What if the bird is a cooked turkey.
> Student 4. "Frozen birds can't fly either."
> Student 5. "Recently hatched baby birds can't fly."

In this silly conversation in an open world of possibilities, exceptions are everywhere. Every new context brings more special cases. Citing Dreyfus and Dreyfus (1986), Stuart Levine calls out the ubiquity of context-dependent knowledge rather than a few simple rules.

> "'… expert-level' performance is associated with a mess of special cases, exceptions, and patterns that people struggle to articulate clearly, and yet can leverage seamlessly in the moment when the situation demands it." (Levine, S., 2021)

Some kinds of domains and problems were never considered. For example, people did not write logical rules for making a good golf swing. They did not create rules for a self-driving cars that could notice nuances of human behavior, such as when a preoccupied parent is trying to manage children at a cross walk. The established approach was to interview experts to determine what knowledge they were using and then to program that knowledge into a computer. A key problem with approaches akin to taking verbal protocols is that experts did not articulate and often lacked conscious awareness of the knowledge that they used.

---

[10] For example, Nilsson's *Principles of Artificial Intelligence* (Nilsson, 1982) aims to connect applications with general methods, focusing on search and logic. Stefik's *Introduction to Knowledge Systems* focuses on representation, search and sample architectures for knowledge system applications built in the 1990s. Russell and Norvig's *Artificial Intelligence: A Modern Approach* unifies AI systems around technologies for creating intelligent agents and is a comprehensive review of multiple waves of AI technology.

A widely held principle and "bitter lesson" from big data machine learning is that large amounts of computation and data about special cases works better than manually engineering knowledge (Sutton, 2019). Restated machine learning scales better in domains that have many special cases. And open world domains *always* have special cases. As Dreyfus and Dreyfus observed long before deep learning and other big data methods were developed, manually built knowledge systems cannot reach expert level performance because they cannot capture all of the exceptional conditions.

Public concern about AI's today concerns its trustworthiness and why it should be more human centered. This is complemented by increasing researcher interest in capabilities that the mainstream approaches to AI do not address. At the same time, the surprising demonstrations of large language models – both in what they seem to do and the ways that they fail – are raising interest in alternatives to the mainstream approaches to AI.

How could AIs be created differently, perhaps having more of the capabilities that people have? A source of inspiration for answering this question is at hand. Since the 1990s, some AI researchers have taken inspiration from developmental psychology. In their first two to three years, human infants develop skills for perceiving the world, learning to move and manipulate objects, learning a native language, developing social skills and exhibiting collaborative behavior. This has led to the creation of a subfield of *developmental robotics* (e.g., (Cangelosi & Schlesinger, 2015)).

Researchers in developmental robotics and embodied systems have been working synergistically with researchers in developmental psychology and neuroscience. As Piaget, an early developmental psychologist observed, young children actively create models of the world. This approach of learning like children do concurs with the mainstream dictum of big data machine learning that robust competence requires enormous amounts of data to develop models. In broad strokes, the required competencies include robust perception, motion, taking effective actions in the world, developing models for how the world works (including other people), and acquiring natural language. What's different is that the data taken in by children (and animals) – and next by developmental AIs – is acquired by sensing and taking action in the world. Piaget called this a sensorimotor approach.

This is the first of two position papers. The second paper proposes a bootstrapping approach to developmental AI (Stefik & Price, 2023). Most human children ultimately develop adult-level intelligence and learn to collaborate. The second paper suggests that developmental robotics research projects stop too soon. What is needed is a sustained bootstrapping approach that considers the competences that children acquire as they transition from newborns to adults.

*Acknowledgments.* Thank you to Robert Hoffman and William Clancey for comments and suggestions on early drafts of this position paper. A special thank you to Ben Shneiderman – a determined advocate for human-centered approaches for his advice on clarifying the message of this paper. He brought my attention to additional literature on the issues and terminology concerning human-robot teaming. Thank you also to many colleagues at the Palo Alto Research Center and to Kai Goebel, director of the Intelligent Systems Laboratory, for patience over several months as I was figuring out next big things to work on.

57. Pirolli, P., Card. S.K. (2005) The sensemaking process and leverage points for analyst technology as identified through cognitive task analysis. *Proceedings of international Conference on Intelligence Analysis.* https://www.researchgate.net/publication/215439203_The_sensemaking_process_and_leverage_points_for_analyst_technology_as_identified_through_cognitive_task_analysis
58. Resnick, L.B., Levine, J.M., & Teasley, S.D. (1996) *Perspectives on Socially Shared Cognition.* American Psychological Association. https://psycnet.apa.org/doiLanding?doi=10.1037%2F10096-000
59. Russell, D.M., Stefik, M.J., Pirolli, P., Card, S.K. (1993) The cost structure of sensemaking, *Proceedings INTERCHI '93 Conference on Human Factors in Computing Systems.* pp. 269-276. https://doi.org/10.1145/169059.169209
60. Russell, S. (2019) *Human Compatible: Artificial Intelligence and the Problem of Control.* Penguin Books, Random house. New York, New York. Viking, imprint of Penguin Random House, New York, New York.
61. Ryan, M. (2018) Human-Machine Teaming for Future Ground Forces. *Center for Strategic and Budgetary Assessments.* https://csbaonline.org/uploads/documents/Human_Machine_Teaming_FinalFormat.pdf
62. Saenz, M.J., Revilla, E., Simón, Cristina. (2020) Designing AI Systems with Human-Machine Teams. *MIT Sloan Management Review.* https://sloanreview.mit.edu/article/designing-ai-systems-with-human-machine-teams/
63. Schmidt, E., Work, B. (2021) *National Security Commission on Artificial Intelligence – Final Report.* https://www.nscai.gov/2021-final-report/
64. Seeber, I., Bittner, E., Briggs, R.O., de Vreede, T., de Vreede, G-J, Elkins, A., Maier, R., Merz, A.B., Oeste-Reiß, S., Randrup, N., Schwabe, G., Söllner, M. (2020) Machines as teammates: A research agenda on AI in team collaboration. *Information & Management* 57. July 2019. https://doi.org/10.1016/j.im.2019.103174
65. Select Committee on Artificial Intelligence, Kratsios (2019) *The National Artificial Intelligence Research and Development Strategic Plan: 2019 Update*, National Science and Technology Council. https://www.nitrd.gov/pubs/National-AI-RD-Strategy-2019.pdf
66. Shneiderman, B. (2022) *Human-Centered AI.* Oxford, Oxford University Press.
67. Smithsonian (2021) Smithsonian FUTURES Teams with Autodesk to Debut Future of Human-A.I. Design. News Release April 21, 2021.
68. https://www.si.edu/newsdesk/releases/smithsonian-futures-teams-autodesk-debut-future-human-ai-design
69. Shrage, M. (1993) The Day You Discover that Your House is Smarter than you Are. *Los Angeles Times.* https://www.latimes.com/archives/la-xpm-1993-11-25-fi-60788-story.html
70. Shyu, H. (2022) Technology Vision for an Era of Competition, News Release - Under Secretary of Defense for Research and Engineering. https://www.cto.mil/wp-content/uploads/2022/02/usdre_strategic_vision_critical_tech_areas.pdf
71. Siddarth, D., Acemoglu, D., Allen, D., Crawford, K., Evans, J., Jordon, M., Weyl, E.G. (2021) How AI Fails Us. *Technology and Democracy Discussion Paper.* https://ethics.harvard.edu/how-ai-fails-us#:~:text=The%20dominant%20vision%20of%20artificial,rather%20than%20social%20and%20relational.
72. Smith, R.P. (2021) What if humans and artificial intelligence teamed up to build better communities? *Smithsonian Magazine.* https://www.smithsonianmag.com/smithsonian-

institution/what-if-humans-and-artificial-intelligence-teamed-build-better-communities-180977541/
73. Stefik, M. (1986) The Next Knowledge Medium. *AI Magazine.* 7(1) pp. 34-67.
74. Stefik, M., Brown, J.S. (1989) Toward Portable Ideas. In Olson, M.H. (Ed.) *Technological Support for Work Group Collaboration*, Hillsdale, New Jersey. Lawrence Erlbaum Associates.
75. Stefik, M.J., Foster, G., Bobrow, D.G., Kahn, K., Lanning, S., Suchman, L. (1987) Beyond the Chalkboard: Computer Support for Collaboration and Problem Solving in Meetings. *Communications of the ACM*, 30(1), pp. 32-47.
76. Stefik, M. (1995) *Introduction to Knowledge Systems.* Morgan Kaufman Publishers, San Francisco.
77. Stefik, M. (1999), "Indistinguishable from Magic: The Real, The Magic, The Virtual" in *The Internet Edge.* The MIT Press, Cambridge, MA. pp. 253-285.
78. Stefik, M., Stefik, B. (2004) *Breakthrough! Stories and Strategies of Radical Innovation.* The MIT Press. Cambridge, MA.
79. Stefik, M., Youngblood, M., Pirolli, P., Lebiere, C., Thomson, R., Price, R., Nelson, L.D., Krivacic, R., Le, J. Mitsopoulos, K., Somers, S., Schooler, M. (2021) Explaining autonomous drones: An XAI journey. *Applied AI Letters*, e54. doi:10.1002/ail2.54
*80.* Stefik, M., Price, R. (2023) *Bootstrapping Developmental AIs: From Simple Competences to Intelligent, Human-Compatible AIs arXiv* https://arxiv.org/abs/2308.04586
81. Sutton, R. (2019) *The bitter lesson.* http://www.incompleteideas.net/IncIdeas/BitterLesson.html
82. Tenenbaum, J.B. (2019), DARPA Artificial Intelligence Colloquium. https://www.youtube.com/watch?v=usRf7fYDgmw
83. Truitt, E.R. (2015) *Medieval Robots: Mechanism, Magic, Nature, and Art (The Middle Ages),* University of Pennsylvania Press.
84. Truitt, E.R. (2021) *Surveillance, Companionship, and Entertainment: The Ancient History of Intelligent Machines.* The MIT Press Reader. https://thereader.mitpress.mit.edu/the-ancient-history-of-intelligent-machines/
85. Veloso, M. (2012) Manuela Veloso's Research Publications. https://www.cs.cmu.edu/~mmv/Veloso.html
86. Vinkhuyzen, E., Cefkin, M. (2016) Developing Socially Acceptable Autonomous Vehicles. *Ethnographic Praxis in Industry Conference.* pp. 522-534. https://anthrosource.onlinelibrary.wiley.com/doi/epdf/10.1111/1559-8918.2016.01108
87. Vinkhuyzen, E. (2022) The trouble with self-driving cars. Erik Vinkhuyzen's video blog. https://www.youtube.com/watch?v=xwif-h4DeLo
88. Waldrop, M. (2001) *The Dream Machine: J.C.R. Licklider and the Dream that Made Computing Personal*, Viking Books, New York, New York.
89. Warneken, F., Tomasello, M. (2006) Altruistic Helping in Human Infants and Young Chimpanzees. *Science* 311, pp. 1301-1303.
90. Wilson, H.J., & Daugherty, P.R. (2018), Collaborative Intelligence: Humans and AI are joining forces. *Harvard Business Review*, July-August, pp. 114–123.
24